\documentclass[conference]{IEEEtran}
\IEEEoverridecommandlockouts
\usepackage{amsmath,graphicx}
\usepackage{amsmath, amsfonts, amssymb}
\usepackage{booktabs} 
\usepackage{multirow} 
\usepackage{cite} 
\usepackage{tabularx}
\usepackage{stackengine}

\usepackage{array}
\usepackage{ragged2e}
\usepackage{caption}
\usepackage{url}
\usepackage{xcolor}
\usepackage{pifont}
 \usepackage{threeparttable}

\definecolor{myred}{RGB}{0,0,0}
\definecolor{red}{RGB}{255,0,0}
\definecolor{myblue}{RGB}{0,0,255}

\newcommand{\pc}{punctuated }
\usepackage{cite}
\usepackage{amsmath,amssymb,amsfonts}
\usepackage{algorithmic}
\usepackage{graphicx}
\usepackage{textcomp}
\usepackage{xcolor}
\def\BibTeX{{\rm B\kern-.05em{\sc i\kern-.025em b}\kern-.08em
    T\kern-.1667em\lower.7ex\hbox{E}\kern-.125emX}}
\begin{document}

\title{End-to-end Joint Punctuated and Normalized  ASR \\with a Limited Amount of Punctuated Training Data
}

\author{
\IEEEauthorblockN{Can Cui\thanks{This work was mostly conducted while Can Cui was pursuing her PhD with Inria Centre at Université de Lorraine (Nancy, France).}}
\IEEEauthorblockA{\textit{iFLYTEK Co., Ltd.} \\
Shanghai, China \\
cancui11@iflytek.com
}
\and
\IEEEauthorblockN{Imran Sheikh}
\IEEEauthorblockA{\textit{Vivoka} \\
Metz, France \\
imran.sheikh@vivoka.com
}
\and
\IEEEauthorblockN{Mostafa Sadeghi, Emmanuel Vincent}
\IEEEauthorblockA{\textit{Université de Lorraine, CNRS, Inria, LORIA, F-54000 } \\
Nancy, France \\
\{mostafa.sadeghi, emmanuel.vincent\}@inria.fr
}

}
\maketitle

\begin{abstract}
Joint punctuated and normalized automatic speech recognition (ASR) aims at outputing transcripts with and without punctuation and casing. This task remains challenging due to the lack of paired speech and punctuated text data in most ASR corpora. We propose two approaches to train an end-to-end joint punctuated and normalized ASR system using limited punctuated data. The first approach uses a language model to convert normalized training transcripts into punctuated transcripts. This achieves a better performance on out-of-domain test data, with up to 17\% relative Punctuation-Case-aware Word Error Rate (PC-WER) reduction. The second approach uses a single decoder conditioned on the type of output. This yields a 42\% relative PC-WER reduction compared to Whisper-base and a 4\% relative (normalized) WER reduction compared to the normalized output of a punctuated-only model. Additionally, our proposed model
demonstrates the feasibility of a joint ASR system using as little as 5\% punctuated training data with a moderate (2.42\% absolute) PC-WER increase.
\end{abstract}


\section{Introduction}
\label{sec:intro}
Transcribing speech into text with punctuation and casing has been an active area of research in automatic speech recognition (ASR)
\cite{fiscus2007rich,klejch2016punctuated,nguyen2019fast}. \textcolor{black}{According to some definitions,\footnote{\url{https://dictionary.cambridge.org/grammar/british-grammar/punctuation}} casing is considered as part of punctuation. For notational simplicity, we therefore refer to punctuated-cased ASR/transcripts as \emph{punctuated} ASR/transcripts in the rest of this paper.}
Joint punctuated and normalized ASR, which produces transcripts both with and without punctuation and casing, is highly desirable because \textcolor{black}{(a) it improves human readability (b) it extends compatibility with natural language processing models that either exploit or discard punctuation information, and (c) it simplifies model deployment and maintenance.}

The conventional \pc ASR approach post-processes normalized ASR output using a punctuation and case restoration model \cite{liao2023improving, huang2021token,polacek23_interspeech}, such as a modified Recurrent Neural Network-Transducer (RNN-T) ASR \cite{ghodsi2020rnn} with a fine-tuned language model (LM) \cite{clark2020electra}. While this avoids \pc transcripts in ASR training, it increases model size, inference time, and lacks acoustic cues for punctuation.  
An alternative is end-to-end (E2E) ASR, which directly generates \pc transcripts \cite{futami2023streaming,tanaka2021end}. A notable example is Whisper \cite{radford2023robust}, trained on large-scale data.
E2E \pc ASR models are less accurate in word recognition than equal-sized normalized ASR models, leading to poorer normalized ASR performance \cite{nozaki2022end,ihori23_interspeech} and increased computation time for output normalization.  
To address this, \cite{nozaki2022end} introduced an auxiliary connectionist temporal classification loss for transcribing normalized text at an intermediate layer, while \cite{ihori23_interspeech} proposed three decoders for spoken, written, and joint transcripts. These methods enable joint punctuated and normalized ASR without increasing model size but require a fully punctuated ASR corpus for training.
Beyond punctuation and casing, \cite{peyser2019improving} introduced a transducer-based ASR with Inverse Text Normalization (ITN), requiring a large training set and longer inference context \cite{peyser2019improving,sunkara2021neural}, leading to higher latency and RTF. In contrast, we aim to develop a low-latency, low-RTF E2E joint \pc and normalized ASR system.

\begin{figure*}[t]
  \centering
  \includegraphics[width=0.9\linewidth]{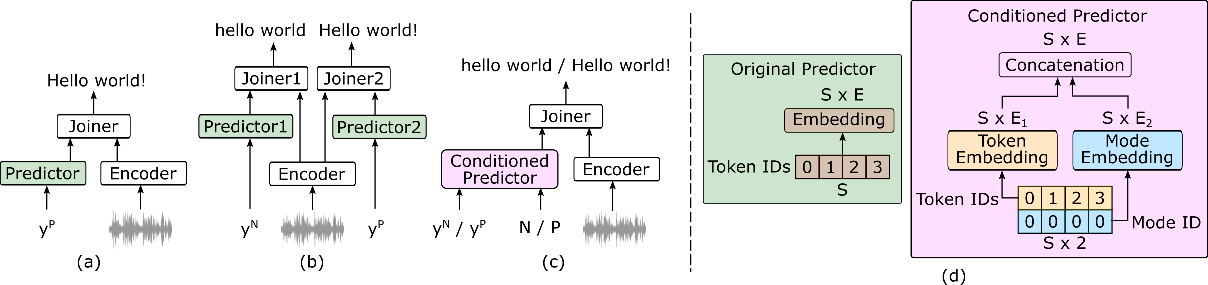}
    \vspace{-5pt}
  \caption{(a) Stateless transducer-based \pc ASR \cite{nozaki2022end}, (b) the proposed 2-Decoder (Joiner + Predictor) joint normalized-\pc ASR, (c) the proposed conditioned Predictor ASR, (d) input layers in the original and conditioned Predictors.  }
  \label{fig:systems}
  \vspace{-15pt}
\end{figure*}


Traditionally, ASR training corpora contain only normalized transcripts, making them unsuitable for \pc ASR models. For example, the 900h CGN corpus \cite{schuurman2003cgn}, a major Dutch ASR resource, lacks \pc data. The 100h Dutch CommonVoice corpus \cite{ardila2019common} includes \pc transcripts but consists of isolated sentences, limiting generalization to long utterances.  
Audiobooks provide punctuated long-form speech, but creating ASR corpora from them is challenging. The Multilingual Librispeech corpus \cite{pratap20_interspeech} has 1,500h of Dutch data but only 40 speakers, and its transcripts are normalized.  
This imbalance between \pc and normalized ASR training data poses a challenge for training an E2E joint \pc and normalized ASR model.

This paper proposes an E2E joint \pc and normalized ASR system that is (a) efficient in both tasks, (b) trainable with limited \pc labeled data, and (c) suitable for streaming. We introduce and compare two complementary approaches to train a stateless transducer-based E2E joint \pc and normalized ASR model. The first approach uses an LM to generate \pc training transcripts.
However, such LMs may not be accurate enough or available for certain domains \cite{sunkara-etal-2020-robust} and/or languages. To address such scenarios, we propose a second approach in which a single decoder is conditioned on the type of output. Experimental results show that our first method results in a 17\% relative error reduction, while the second method enables training with an exceptionally low proportion of \pc data.

The paper is organized as follows. 
Section~\ref{sec:preliminaries} presents the preliminaries.  Section~\ref{sec:proposed-method} introduces our approaches.
Section~\ref{sec:experiments} describes our experiments. We conclude in Section~\ref{sec:conclusion}.


\section{\textcolor{myred}{Preliminary of stateless \\ transducer-based \pc ASR}}
\label{sec:preliminaries}
An E2E \pc ASR system \cite{nozaki2022end} directly transcribes speech into a \pc transcript. 
Given an acoustic feature sequence \(X \in \mathbb{R}^{L\times A} \) where  $L$ is the sequence length and $A$ the feature dimension, the training objective is to maximize the probability
\vspace{-10pt}
\begin{equation}\label{eq:eq1}
P(Y^\text{P}|X) = \prod_{s=1}^{S^\text{P}} P(y^\text{P}_s|{y}^\text{P}_{[1:s-1]},X)
\vspace{-5pt}
\end{equation}

by generating a sequence \(Y^\text{P} \in \mathbb{R}^{S^\text{P}} \), where ``P'' stands for \pc transcription, and $S^\text{P}$ represents the \pc sequence length. The loss function can be written as
\begin{equation}\label{eq:eq2}
\mathcal{L}^\text{P} = - \sum_{Y^\text{P},X} \log P(Y^\text{P}|X).
\end{equation}
A stateless transducer-based E2E ASR system, which uses an RNN-T framework \cite{graves2012sequence} with a stateless prediction network \cite{ghodsi2020rnn}, can be readily extended to the \pc transcription task (see Figure~\ref{fig:systems}(a)). \textcolor{myred}{Zipformer \cite{yao2023zipformer}, a stateless transformer encoder with downsampling and a zip-like structure, further enhances efficiency while maintaining accuracy. } This brings the natural streaming recognition capability of RNN-T to \pc ASR.


\section{Proposed methods}
\label{sec:proposed-method}

\subsection{2-Decoder joint normalized-\pc ASR}
\label{subsec:2-dec}
Inspired by \cite{nozaki2022end, ihori23_interspeech}, we design an ASR system with two Decoders, each consisting of its own Predictor and Joiner (see Figure~\ref{fig:systems}(b)).
Using the output of the same Encoder,
these two Decoders generate 
the
\pc and normalized transcripts. Their respective training objectives are\footnote{These equations are kept consistent with (1) and do not account for the stateless and/or streaming operation of the transducer model.}
\begin{align}
\vspace{-5pt}
P(Y^\text{N}|X) &= \prod_{s=1}^{S^\text{N}} P({y}^\text{N}_{s}|{y}^\text{N}_{[1:s-1]},X),\label{eq:eq3}\\
P(Y^\text{P}|X) &= \prod_{s=1}^{S^\text{P}} P({y}^\text{P}_{s}|{y}^\text{P}_{[1:s-1]},X),\label{eq:eq4}
\vspace{-5pt}
\end{align}
where ``N'' stands for normalized transcription, and $S^\text{N}$ represents the normalized sequence length. The joint loss function is defined as
\begin{align}
\mathcal{L}^\text{2-decoder} &= \mathcal{L}^\text{N} + \mathcal{L}^\text{P} \label{eq:eq5}\\
&= - \sum_{Y^\text{N},X} \log P(Y^\text{N}|X) - \sum_{Y^\text{P},X}\log  P(Y^\text{P}|X).\label{eq:eq5-2}
\end{align}

\subsection{\textcolor{black}{Training ASR using auto-\pc transcripts}}
\label{sec:auto-punctuated-cased}

\textcolor{black}{Automatically generated \pc transcripts of ASR training data can be used to train \pc ASR models in the absence of human-generated \pc transcripts. To the best of our knowledge, the use of auto-\pc transcripts to train \pc ASR models has been unexplored in prior works. The work in \cite{tanaka2021end} proposed a semi-supervised rich ASR training method in which a rich ASR system is first trained on a small amount of human-labeled rich training data and then used to automatically generate auto-rich transcripts of a larger unlabeled speech corpus. This approach focused on the rich transcription of speech phenomena such as fillers, laughter, coughs, etc., hence some human-labeled data is necessary. By contrast, our transcription task focuses on punctuation and casing. Punctuation- and case-enhanced transcripts can be directly obtained from normalized transcripts using state-of-the-art punctuation and case restoration models \cite{liao2023improving, huang2021token,polacek23_interspeech}. Interestingly, this approach can be applied in scenarios where there is no \pc labeled training data at all. }

\subsection{Conditioned Predictor ASR}
A drawback of the auto-\pc transcription-driven ASR training approach presented above is that errors made by the punctuation and case prediction model may propagate into the \pc ASR model. Moreover, such case and punctuation restoration models may not be accurate enough, or even available, for certain domains \cite{sunkara-etal-2020-robust} and/or languages. Hence, we propose a second training approach to address such scenarios. 

Our approach effectively utilizes the small amount of \pc training data and the large amount of normalized training data by using a single conditioned Predictor to handle both \pc and normalized transcriptions.
As shown in Figure~\ref{fig:systems}(c), the Predictor is conditioned on the transcription mode ID, which is an input that specifies whether we want a normalized (N) or \pc (P) output. The mode ID input is inspired by the language ID information used by E2E multilingual ASR models\cite{li2022recent, toshniwal2018multilingual}. As shown in Figure~\ref{fig:systems} (d), the token embeddings are concatenated with the mode embedding before feeding them to the following Predictor layers.
The Conditioned Predictor ASR system can use a loss function similar to Equation \eqref{eq:eq4}. 
However, only part of the input samples will have a \pc reference transcript and will use the corresponding loss.
\textcolor{black}{This could lead to uneven model performance between the two output modes, particularly poor performance of the \pc transcription mode. To address this issue, we employ a tradeoff parameter $\alpha$ to adjust the weights of $\mathcal{L}^\text{N}$  and $\mathcal{L}^\text{P}$, as follows:}
\begin{equation}\label{eq:eq6}
\mathcal{L}^\text{cond-predictor} =  (1-\alpha)\mathcal{L}^\text{N} + \alpha \mathcal{L}^\text{P}.
\end{equation}

\section{Experiments}
\label{sec:experiments}
This section details our experimental setup and results. For reproducibility, our code is available online.\footnote{\url{https://github.com/can-cui/punctuated-normalized-asr}}

\subsection{Dataset and metrics}
\label{subsec:data}
\subsubsection{Dataset}
\label{subsubsec:data}
We conduct experiments on LibriSpeech \cite{panayotov2015librispeech}, using train-960 for training, dev-clean/dev-other for validation, and test-clean/test-other for testing. We retrieve the \pc transcripts of LibriSpeech from the original Project Gutenberg texts.\footnote{Available at \url{https://www.openslr.org/12/}} \textcolor{black}{Erroneous sentences, such as fully uppercase chapter names, are removed. A recent work on building a \pc Librispeech corpus \cite{meister2023librispeech} has detailed retrieval steps similar to those we have developed independently}. To evaluate the model's performance on real unseen data, we use the \textcolor{myred}{CommonVoice \cite{commonvoice:2020} English clean test set} and independent headset microphone recordings of the AMI meeting corpus \cite{carletta2005ami} test set along with their original \pc transcripts.

\begin{table*}[t]
\caption{Error rates (\%) \textcolor{myred}{on in-domain{\ding{61}} test data} for all the ASR models trained on LibriSpeech train-960 with different types of transcripts (Trans): normalized (N), original \pc (P), or auto-\pc (P\textquotesingle). All the test ground truth transcripts are the original \pc transcripts.
The 2-Decoder ASR and the Conditioned Predictor ASR systems use both \pc (P or P\textquotesingle) and normalized (N) transcripts for the entire train-960 set.
}
\vspace{-5pt}
\centering
\label{table:all-asr}
\scalebox{0.95}{
\addtolength{\tabcolsep}{-0.2em}
\begin{tabular}{c|c|c|c|cccc|cccc|cccc|c}
    \toprule
    \multirow{2}{*}{\bfseries Config ID} & 
    \multirow{2}{*}{\bfseries Type of ASR} & 
    \multirow{2}{*}{\bfseries \# Param} & 
    \multirow{2}{*}{\bfseries Trans} & 
    \multicolumn{4}{c|}{\bfseries LibriSpeech test-clean} & 
    \multicolumn{4}{c|}{\bfseries LibriSpeech test-other}& 
    \multirow{2}{*}{\bfseries RTF} 
    \\ 
    \cmidrule(lr){5-8} \cmidrule(lr){9-12} 
    &&&& \textbf{WER} &\textbf{PuncER}& \textbf{CaseER}& \textbf{PC-WER} & \textbf{WER} &\textbf{PuncER}& \textbf{CaseER}& \textbf{PC-WER}
    \\ 
    \cmidrule(lr){1-13}
   0& \textcolor{myred}{ Whisper-base} & 74 M&-&5.80&33.86&32.04&15.96&13.02&36.79&27.72&19.48&\textbf{0.09}\\
    \hline
   1& \textcolor{myred}{ Whisper-small} & 244 M&-&4.40&31.19&28.49&10.95&11.89&32.17&25.67&17.59&0.23\\
    \hline
    2&Cascaded (ASR+LM) & 180 M & N &\textbf{2.18 }&44.33&35.60&11.70&\textbf{5.03}&47.69&34.82&14.91& 0.79\\
    \hline
    3&\multirow{2}{*} {Punctuated only (a)} & 70 M &P&2.45&\textbf{30.39}&28.21&\textbf{9.12}&5.73&30.51&\textbf{25.14}&11.94&0.45\\
    4&& 70 M &P\textquotesingle &2.35&44.43&31.42&11.60	&5.49&47.39&30.14&14.94&0.44\\
    \hline
    5&\multirow{2}{*} {2-Decoder (b)} & 71 M &P + N&\textbf{2.33}&\textbf{29.99}&\textbf{26.99}&	\textbf{8.85}	&5.45	&31.05&\textbf{24.76}&	\textbf{11.76}&0.58\\
   6& & 71 M &P\textquotesingle  + N&\textbf{2.28}&45.10	&31.68&11.67&\textbf{5.38}&47.81&29.53&14.88&0.59\\
    \hline
    7&\multirow{2}{*} {Cond Predictor (c)} & 70 M &P + N &\textbf{2.35}&\textbf{29.14}&35.61&\textbf{9.32}&5.49&\textbf{29.62}&30.77&12.01&0.44\\
    8&& 70 M &P\textquotesingle  + N&\textbf{2.25}&44.60&41.86	& 12.30&\textbf{5.24}&46.84&33.99&15.42& 0.45\\

    \bottomrule
\end{tabular}
}
\begin{tablenotes}
\centering
      \footnotesize
      \item \ding{61}: \textcolor{myred}{For Whisper, it is unknown whether the test data is in-domain or out-of-domain due to the lack of training data information.}
      \item \textcolor{black}{Note: We employed the SCTK toolkit \cite{sctk} to conduct Matched Pair Sentence Segment statistical significance tests adapted to the four metrics. In all the tables, we highlight in bold the best result in each column and the results statistically equivalent to it at a 0.05 significance level.}
    \end{tablenotes}
    \vspace{-10pt}
\end{table*}

    

\subsubsection{Metrics}
\label{subsubsec:metrics}

The standard Word Error Rate (WER) is insufficient for evaluating \pc transcription. Traditionally, Precision, Recall, and F1 scores have been used \cite{huang2021token, polacek23_interspeech}, but they report separate scores for punctuation marks and lack a standardized casing metric.  
\textcolor{myred}{Recent work \cite{meister2023librispeech} proposed adapted error rate metrics for punctuation and case, underscoring the need for improved evaluation. However, case error rates are missing, and punctuation error rates include word errors, failing to isolate punctuation-specific mistakes.}  
To better assess punctuation and casing performance and standardize error comparisons, we introduce Punctuation-only Error Rate (PuncER), Case-only Error Rate (CaseER) and Punctuation-Case-aware WER (PC-WER) alongside the WER. These metrics are calculated as:
\begin{align}
\text{PuncER} &= \frac{E_{\text{p-nc}} - E_{\text{np-nc}}}{N_{\text{p}}} , & \text{CaseER} &= \frac{E_{\text{np-c}} - E_{\text{np-nc}}}{N_{\text{c}}},\nonumber\\
\text{PC-WER} &= \frac{E_{\text{p-c}}}{N_{\text{p-c}}} , & \text{WER} &= \frac{E_{\text{np-nc}}}{N_{\text{np-nc}}},\label{eq:wer}
\end{align}
where $E$ represents the total number of substitution, deletion, and insertion errors, and $N$ represents the total number of words, punctuation and/or casing marks. 
Suffixes ``p'' and ``c'' stand for the presence of punctuation and casing whereas suffices ``np'' and ``nc'' stand for the absence of punctuation and casing, respectively. 

To evaluate inference speed in streaming, we use Real-Time Factor (RTF), defined as inference time divided by audio length. For Whisper models, we load them once and process test audio sequentially. For transducer-based models, we export to ONNX and follow the same process. RTF is measured during CPU-based inference for punctuated transcription.

\subsection{Model and training setup}
We use 80-dimensional log Mel filterbank features as input for all ASR models and a SentencePiece tokenizer \cite{kudo2018sentencepiece} with a 500-word vocabulary.  
\textcolor{myred}{Pretrained Whisper models \cite{radford2023robust} serve as baselines: Whisper-base (74M parameters) for comparability and Whisper-small (244M) for high-performance reference.}  
Our ASR models (E2E ASR, 2-Decoder ASR, Conditioned Predictor ASR) follow the pruned stateless-transducer recipe with a Zipformer encoder in icefall.\footnote{\url{https://github.com/k2-fsa/icefall/blob/master/egs/librispeech/ASR/pruned_transducer_stateless7}}  
For Conditioned Predictor ASR, Token Embedding \(E_1\) and Mode Embedding \(E_2\) have 500 and 12 dimensions, respectively. All models were trained for 40 epochs on 4 Nvidia RTX 2080 Ti GPUs, with inference on an Intel Xeon Gold 5218R CPU.

We use the Rpunct \cite{daulet} model to generate \pc transcripts from normalized transcripts. This model is derived from the BERT \cite{kenton2019bert} LM after fine-tuning it for English punctuation and case restoration tasks. 


\subsection{Evaluation results}
\subsubsection{Effectiveness of E2E models}
Table~\ref{table:all-asr} presents the error rates \textcolor{myred}{on in-domain test sets of our ASR models}. \textcolor{myred}{First of all, the transducer-based ASR models demonstrate superior performance compared to Whisper models. Specially, our proposed Conditioned Predictor ASR achieves a PC-WER reduction of up to 42\% relative (from 15.96\% to 9.32\%) compared to Whisper-base, which is of comparable size, and 15\% relative (from 10.95\% to 9.32\%) compared to Whisper-small. For the transducer-based ASR models,} the conventional cascaded system, which uses a normalized output ASR model followed by a punctuation and case restoration LM, is included as a baseline. Comparison of error rates of the cascaded system and E2E models trained using \pc data shows that the latter can lead to significant reductions in PC-WER. For instance, when using the 2-Decoder ASR model (P + N) we obtain a relative {PC-WER} reduction of up to 24\% (from 11.70\% to 8.85\%). 

When comparing the 2-Decoder ASR model with the proposed Conditioned Predictor ASR model on the LibriSpeech test set, the latter exhibits a decrease in PuncER. For example, when using the same P + N training set, the Conditioned Predictor ASR model achieves a relative decrease of 3\% on test-clean and 5\% on test-other in PuncER compared to the former. But 2-Decoder ASR has a better performance in terms of case generation, by reducing the CaseER by 24\% relative on test-clean (from 35.60\% to 26.99\%). 

\textcolor{myred}{Furthermore, the proposed Conditioned Predictor ASR gives both normalized and punctuated outputs. The normalized output shows a 4\% relative reduction in WER on test-clean (from 2.45\% to 2.35\%) and test-other (from 5.73\% to 5.49\%). This demonstrates that using a dual-output ASR model yields better performance for normalized output compared to normalizing a Punctuated-only ASR model.}

\textcolor{myred}{
Additionally, from the perspective of inference speed, all transducer-based ASR models achieve an RTF lower than 1, although they are higher than Whisper models. Notably, the proposed Conditioned Predictor ASR model has the lowest RTF, which is 0.35 lower (from 0.79 to 0.44) than the traditional cascaded system and 0.15 lower (from 0.59 to 0.44) than the 2-Decoder model, while maintaining similar performance. The Punctuated-only model has an RTF comparable to the Conditioned Predictor model, but this value will increase if normalization of the ASR output is required.}

\subsubsection{Original vs.\ auto-\pc transcripts for training}
\textcolor{myred}{Table~\ref{table:all-asr-unseen} displays the test results on the CommonVoice and the AMI test sets. On out-of-domain test sets, our transducer-based models still outperform Whisper models, with a relative lower PC-WER up to 38\% (from 73.27\% to 45.24\%).}

Comparing Table~\ref{table:all-asr} and Table~\ref{table:all-asr-unseen} reveals key insights on original vs. auto-punctuated training data. E2E ASR trained on original \pc transcripts achieves lower punctuation and case error rates on in-domain LibriSpeech test sets, while using auto-\pc training transcripts improves performance on out-of-domain CommonVoice and AMI test sets.  
For example, the 2-Decoder ASR model shows a 32\% relative PC-WER increase (from 8.85\% to 11.67\%) on test-clean and 27\% (from 11.76\% to 14.88\%) on test-other when using auto-\pc training transcripts (P\textquotesingle+N) instead of original transcripts (P+N).

\begin{table}[t]
\caption{Error rates (\%) on out-of-domain test sets for all the ASR models. The ID column corresponds to the config ID in Table~\ref{table:all-asr}, which specifies the ASR type and transcription used for training.
}
\vspace{-5pt}
\centering
\label{table:all-asr-unseen}
\scalebox{0.93}{
\addtolength{\tabcolsep}{-0.5em}
\begin{tabular}{c|cccc|cccc}
    \toprule
    \multirow{2}{*}{\bfseries ID} & 
    \multicolumn{4}{c|}{\bfseries CommonVoice test-clean} & 
    \multicolumn{4}{c}{\bfseries AMI test}
    \\ 
    \cmidrule(lr){2-5} \cmidrule(lr){6-9} 
    & \textbf{WER} &\textbf{PuncER}& \textbf{CaseER}& \textbf{PC-WER} & \textbf{WER} &\textbf{PuncER}& \textbf{CaseER}& \textbf{PC-WER}
    \\ 
    \cmidrule(lr){1-9}
    0&38.06&34.03&11.09&38.99& 71.72&64.22&39.84&73.27\\
    \hline
    1&37.90&36.72&\textbf{10.56}&39.15&58.49&52.99&36.48&60.12\\
    \hline
    2&\textbf{27.90}&\textbf{27.91}&11.63&\textbf{29.50}&\textbf{37.46}&74.84&45.92&\textbf{45.99} \\
    \hline
    3&29.23&64.87&16.47&36.61&39.59&97.02&38.57&50.61\\
     4&\textbf{27.81}&\textbf{27.28}&11.42&\textbf{29.27}&39.01&71.82&37.39&46.33\\
    \hline
    5&28.44&64.35&16.04&35.79	&39.12	&96.48& 39.44&50.18\\
    6&28.35&\textbf{27.65}&\textbf{10.83}&\textbf{29.70}&39.33&\textbf{65.63}&\textbf{33.64}&\textbf{45.43}\\
    \hline
    7&28.57&60.35&23.21&36.35&40.70&96.40&38.40&51.47 \\
    8&\textbf{27.84}&\textbf{27.44}&11.34&\textbf{29.65}&38.77&65.65&35.62&\textbf{45.24} \\

    \bottomrule
\end{tabular}
}
\vspace{-10pt}
\end{table}

However, the same model exhibits a PC-WER reduction of \textcolor{myred}{17\% relative (from 35.79\% to 29.70\%) on the CommonVoice test set and}
9\% relative (from 50.18\% to 45.43\%) on the AMI test set when using auto-\pc training transcripts. 
\textcolor{myred}{The differences mainly stem from the addition of case and punctuation. Models trained with auto-punctuated transcripts show a lower PuncER and CaseER. Specifically, for the Conditioned Predictor model on the CommonVoice test set, the PuncER reduction can be as high as 54\% relative (from 60.35\% to 27.44\%), and the CaseER can be reduced by up to 51\% relative (from 23.21\% to 11.34\%) when using auto-punctuated transcripts.}
\textcolor{black}{This could be explained by the inconsistent (and sometimes erroneous) punctuation in LibriSpeech ground truth punctuated transcripts. By contrast, being trained on a wide range of domains, the LM outputs a better (possibly domain-dependent) punctuation than that learned on the specific LibriSpeech domain on average.}

\subsubsection{Conditioned Predictor ASR with limited \pc data}

Unlike the 2-Decoder ASR model, the Conditioned Predictor ASR model can be trained with different proportions of \pc and normalized training data. To evaluate the effectiveness of this capability, we trained the Conditioned Predictor ASR model with different proportions of \pc and normalized (P + N) training data. Table~\ref{table:punctuated-cased-data} presents the error rates obtained on the LibriSpeech test-clean set. It can be observed that the PC-WER and PuncER increase by small margins even when the amount of \pc training data is reduced drastically. Using only 5\% of \pc training data can still achieve a PC-WER of 11.46\%, i.e., a 2.42\% absolute increase. 
{This shows that the Predictor-Conditioned ASR model learns to generalize punctuation patterns from limited examples by leveraging normalized text and acoustic-text alignment.}
Therefore, this model can serve scenarios having limited or severely limited \pc training data.

\begin{table}[t]
\caption{Error rates (\%) of Conditioned Predictor ASR (c) on LibriSpeech test-clean when using different proportions of \pc and normalized (P + N) training data.
}
\vspace{-5pt}
\centering
\label{table:punctuated-cased-data}
\scalebox{0.95}{
\addtolength{\tabcolsep}{-0.2em}
\begin{tabular}{cccccccc}
    \toprule
    \bfseries (P : N) Proportion & 
    \bfseries WER & 
    
    \bfseries PuncER & 
    \bfseries CaseER & 
    \bfseries PC-WER &
    \\
    \cmidrule(lr){1-5}
    
    0.50 : 0.50 & \textbf{2.38}&29.33&\textbf{34.86}&\textbf{9.32} \\
    0.35 : 0.65 & \textbf{2.37}&\textbf{28.99}&45.44&9.95 \\
     0.20 : 0.80&\textbf{2.34}&29.96&56.37&10.83 \\
    0.05 : 0.95 & 2.56&31.59&58.58&11.46\\
    \bottomrule
\end{tabular}
}
\vspace{-10pt}
\end{table}

\section{Conclusion}
\label{sec:conclusion}
\noindent In this paper, we introduced two approaches for training a stateless transducer-based E2E joint \pc and normalized ASR model with minimal \pc labeled data. The first approach leverages a language model to generate auto-\pc transcripts, achieving up to 17\% relative PC-WER reduction on out-of-domain data compared to training on \pc data. The second approach uses a Conditioned Predictor, enabling efficient dual-output ASR.  
Our Conditioned Predictor model reduces PC-WER by 42\% relative to Whisper-base with a similar parameter count. 
This approach proves feasible with as little as 5\% \pc training data, incurring only a 2.42\% absolute error increase. 


\bibliographystyle{IEEEbib}
\bibliography{strings}
\end{document}